# LesionSpeak: Segmentation-guided LoRA adaptation of a vision-language model for grounded brain MRI report generation

Almustapha A. Wakili[a,*], Muhammad Adam[a], Julius Akan[a], Stephanie Y. Jo[b], Lei Zhang[a], Woosub Jung[a]

[a] *Towson University, Towson, MD, United States*
[b] *Baltimore VA Medical Center and University of Maryland School of Medicine, Department of Biochemistry and Molecular Biology, Baltimore, Maryland, United States*

## ARTICLE INFO

## ABSTRACT

Brain MRI report generation requires more than fluent text. A useful model must identify the lesion, use the MRI sequence context appropriately, and describe findings that match the input image. However, vision-language models trained only on image-report pairs often produce plausible reports that still confuse lesion side, misidentify sequence type, or mention unsupported findings. We present LesionSpeak, a segmentation-guided LoRA adaptation framework that uses lesion masks as training-time supervision without requiring masks at inference. Using 1,938 training MRI–report records from the AutoRG-Brain RadGenome-Brain MRI dataset, we encode segmentation-derived laterality and lesion attributes as structured training supervision and study contour-guided training. In the auxiliary comparison, all compared models receive MRI montages and sequence metadata; the auxiliary lesion-aware vision-adapted variant additionally predicts lesion attributes before generating a report. On 580 test records, without segmentation masks for input construction or generation, the fine-tuned baseline with seed 42 achieves BLEU-1 of 27.80, while the auxiliary variant achieves BLEU-1 of 30.89 ± 1.16, BLEU-4 of 10.46 ± 0.74, ROUGE-L of 34.00 ± 0.95, and BERTScore-F1 of 57.27 ± 0.59 across three training seeds (mean ± sample standard deviation). Reference-text laterality agreement is 70.48 ± 2.59% across three training seeds on 402 eligible test records, compared with 56.22% for the fine-tuned baseline (seed 42). Replacing the patient MRI with another patient's image from the same cohort and MRI sequence reduces laterality agreement by 16.92-24.13 percentage points across the three auxiliary-model seeds, whereas the fine-tuned baseline (seed 42) shows a 0.00-point difference, supporting image dependence of the auxiliary model's reference-text laterality agreement.

## 1. Introduction

Radiology reports are the main way imaging findings are communicated to clinicians and used in patient care (Wallis & McCoubrie, 2011). In brain MRI, a report must do more than describe what appears in an image. It must explain where a lesion is located, how it appears across MRI sequences, whether there is surrounding edema or mass effect, and which findings matter for diagnosis, treatment planning, and follow-up. Assisting this process with AI is therefore useful, but it is also difficult. A useful system must generate fluent text while remaining faithful to the underlying lesion.

*Corresponding author.
*E-mail addresses:* awakili@towson.edu (A.A. Wakili), madam1@students.towson.edu (M. Adam), jakan1@students.towson.edu (J. Akan), stephanie.jo@som.umaryland.edu (S.Y. Jo), leizhang@towson.edu (L. Zhang), woosubjung@towson.edu (W. Jung).

Despite growing interest in automated radiology reporting, systems developed for settings such as chest X-ray have not translated cleanly to brain MRI, where multi-sequence acquisition, volumetric data, and lesion-specific language create distinct challenges (Sloan et al., 2024). More recently, vision-language models (VLMs) have shown strong performance on tasks that combine images and text (Wu et al., 2025; Yang et al., 2024). A representative example is LLaVA, which links an image encoder to a large language model so that the system can answer questions or generate text from visual input (Liu et al., 2024). These models are promising for medical reporting, but brain MRI remains especially demanding. First, brain MRI is usually multi-sequence, meaning that the same case may include T1, T2, FLAIR, and post-contrast images, each showing different tissue properties. Second, MRI is volumetric, while many current VLMs operate most naturally on 2D inputs (Elboardy et al., 2025; Wang et al., 2024). Third, fluent output does not guarantee grounded output. A model may produce plausible radiology language while still confusing lesion side, sequence type, or lesion composition (Lei et al., 2024; Tanida et al., 2023). As a result, strong zero-shot medical VLMs often remain far from expert-style brain MRI reporting (Elboardy et al., 2025; Lei et al., 2024).

In this work, we address that problem through LesionSpeak, a segmentation-guided framework for grounded brain MRI report generation. Our approach builds on LLaVA-v1.5-7B (Liu et al., 2024) and adapts it using low-rank adaptation (LoRA) (Hu et al., 2022), a parameter-efficient method that learns low-rank weight updates. The plain fine-tuning baseline (FT) learns from MRI-report pairs without additional lesion supervision. We train on 1,938 brain MRI-report records derived from the AutoRG-Brain RadGenome-Brain MRI resource (Lei et al., 2024). During training, lesion information from segmentation masks is used as structured hints or auxiliary supervision describing laterality and lesion compartments. MRI sequence identity is acquisition metadata and is treated separately from mask-derived attributes. An auxiliary variant combines lesion-attribute prediction with vision-encoder LoRA. We use contour overlays as training-time spatial cues and evaluate generated reports using text-similarity metrics, automated agreement with reference-text laterality, and image-shuffling controls. These measurements assess text agreement and image dependence rather than clinical accuracy. Throughout this paper, report grounding is operationalized through automated reference-text laterality agreement and its dependence on patient-specific image information.

A central design choice in this study is to separate training-time supervision from inference-time availability. At inference, when the trained model generates a report, input montages are constructed from MRI intensities without segmentation masks. Segmentation-derived hints and contours are withheld in the primary test conditions. In the auxiliary experiments, all compared models receive sequence metadata, and the auxiliary variant generates its own lesion-attribute answers before producing the report. We also evaluate a hint-available reference setting in which the hint remains available at inference. This distinction is clinically important: while lesion masks are often generated during research dataset curation or treatment-planning workflows, they cannot be assumed at inference in routine practice. The experiments further show that the form of lesion supervision matters.

The main contributions of this work are as follows:

(1) We propose LesionSpeak, a lightweight framework for grounded brain MRI report generation that converts segmentation-derived lesion attributes into structured training-time supervision. MRI-only montage construction allows reports to be generated without segmentation masks at test time, while mask-derived attributes provide additional training supervision. (2) We compare focused and full structured hints under explicit inference-time information constraints and examine an auxiliary variant combining lesion-attribute prediction, vision-encoder LoRA, and staged report generation.

(3) We evaluate contour overlays alone and in combination with sequence hints across three training seeds, with an additional single-seed reflected-contour control to assess the effect of spatial correspondence.

(4) We complement text-similarity metrics with automated reference-text laterality agreement, output-diversity measures, and image-shuffling controls. Across three training seeds, the auxiliary variant's laterality agreement decreases when its input is replaced with another patient's MRI of the same cohort and sequence, supporting image dependence of this measured endpoint.

The remainder of the paper is organized as follows. Section 2 reviews related work. Section 3 describes the proposed framework and study design. Section 4 presents the empirical results. Section 5 discusses the main implications and limitations.

## 2. Related work

### 2.1. Medical vision-language models

Recent progress in vision-language models has enabled multimodal clinical reasoning, visual question answering, and report generation across medical imaging tasks. LLaVA introduced a simple architecture that connects a pretrained visual encoder to a large language model through visual instruction tuning (Liu et al., 2023), and LLaVA-1.5 strengthened this framework through improved data curation and a stronger visual backbone (Liu et al., 2024). Medical adaptations extended this idea to biomedical data through domain-specific pretraining and instruction tuning. LLaVA-Med specialized the LLaVA framework using PubMed-derived figure-caption pairs and GPT-assisted instruction generation (Li et al., 2023), while Med-Flamingo showed that few-shot multimodal adaptation can improve medical visual question answering without extensive retraining (Moor et al., 2023). Subsequent work further scaled medical VLMs, including CheXagent, Med-Gemini, and RadFM (Chen et al., 2024; Wu et al., 2025; Yang et al., 2024). Despite this progress, brain MRI remains challenging because it is multi-sequence, often volumetric, and requires lesion-level anatomical precision. Wang et al. (2024) further showed through Vote-MI that lesion-guided 2D slice selection substantially improves VLM performance on volumetric brain MRI compared with random slice selection.

### 2.2. Parameter-efficient adaptation of vision-language models

As multimodal foundation models grow larger, parameter-efficient adaptation has become increasingly important for medical applications. LoRA introduces low-rank trainable updates while freezing the pretrained backbone, reducing trainable parameters and memory cost without changing the inference-time architecture (Hu et al., 2022). This is particularly attractive in medical imaging, where task-specific datasets are much smaller than pretraining corpora and full-model adaptation is often impractical. Hoque et al. (2025) showed that targeted LoRA applied to fewer than 1% of model parameters outperformed broader adaptation strategies for radiology image captioning. Vahdati et al. (2025) likewise identified parameter-efficient tuning as a practical route for radiology-oriented large language models under data and hardware constraints. A lightweight radiology-specific end-to-end VLM recipe has also been explored in chest radiography through LLaVA-Rad, which combines a radiology-oriented visual encoder, a 7B language model, and LoRA-based adaptation for report generation (Chaves et al., 2024). This provides a useful architectural precedent for lightweight domain-specific VLM adaptation, even though the target modality differs from brain MRI. These findings motivate our use of lightweight LoRA adaptation rather than full retraining for brain MRI report generation.

### 2.3. Brain MRI report generation

Brain MRI report generation remains less mature than broader medical multimodal reasoning. Earlier report generation systems such as R2Gen improved long-form radiology report generation for chest X-rays (Chen et al., 2020), but did not address multi-sequence volumetric brain MRI, where lesion location, edema, mass effect, and anatomical involvement must be described with much finer spatial precision. AutoRG-Brain is a key step toward this goal, proposing a grounded brain MRI reporting framework that links report generation to localized visual evidence through anomaly and brain-structure prompts (Lei et al., 2024). The study also introduces RadGenome-Brain MRI, a grounded report-generation dataset containing 3408 imaging-report pairs across five disease categories and six MRI sequence categories. Elboardy et al. (2025) provided a benchmark study for brain cancer diagnosis using multisequence MRI and showed that larger or more general models do not necessarily produce better clinical reporting. More recently, BTReport proposed a brain-tumor reporting framework based on deterministic extraction of clinically relevant features followed by language-model synthesis, showing that structured clinical feature extraction can serve as an effective intermediary between imaging and free-text generation (Rivera et al., 2026). Together, these studies suggest that grounded brain MRI reporting requires more than generic multimodal fluency.

### 2.4. Segmentation-guided medical reporting

Related work in brain MRI has also investigated segmentation frameworks for lesion analysis, including hybrid CNN–Transformer models and federated learning frameworks (Wakili et al., 2026) (Wakili et al., 2025). A related line of work improves medical language generation by incorporating structured lesion or region-level guidance. In chest radiology, Tanida et al. (2023) proposed an interactive and explainable region-guided report generation framework that grounds generated sentences in localized image regions. In brain MRI, AutoRG-Brain provides the clearest example of segmentation-guided reporting because its report generator is conditioned on anomaly and brain-structure prompts rather than relying on unconstrained global generation (Lei et al., 2024). Related work in spinal MRI has further shown that segmentation-aware multimodal supervision improves report generation: Helmy et al. (2026) used segmentation-guided processing and integration of multimodal spinal MRI data for report generation, while Helmy et al. (2025) demonstrated that 3D vision-language models with segmentation-guided multimodal data can improve spinal MRI report quality. Unlike AutoRG-Brain, which conditions its report generator on segmentation prompts during both training and inference, LesionSpeak uses segmentation-derived lesion supervision during training without providing ground-truth lesion annotations at test time. This allows us to study whether training-time lesion supervision affects report generation when ground-truth lesion annotations are unavailable during testing.

## 3. Methods

We build the LesionSpeak framework from AutoRG-Brain-derived MRI-report pairs through MRI-based multislice preprocessing, segmentation-derived training supervision, and LoRA adaptation of LLaVA-v1.5. Fig. 1 summarizes the overall workflow. Segmenta-tion masks provide laterality and lesion-compartment supervision during training, while MRI sequence identity is obtained separately from acquisition metadata. Hint-comparison experiments use structured hints and optional contour overlays, whereas auxiliary experiments use lesion attributes as prediction targets. At inference, we construct test montages without segmentation masks, and we do not provide ground-truth hints, contours, or auxiliary answers.

### 3.1. Dataset preparation and MRI preprocessing

We constructed a brain MRI report-generation corpus from the RadGenome-Brain MRI component of AutoRG-Brain (Lei et al., 2024). The master inventory contains 2828 sequence-level entries from four pathology groups: glioma (GLI), meningioma (MEN), metastasis (MET), and white matter hyperintensity (WMH). Twenty WMH native-T1 entries without sequence-specific reference findings were excluded, leaving 2808 eligible MRI-report records. All records from the same patient are assigned to a single partition, giving training, validation, and test sets of 1938, 290, and 580 records from 483, 75, and 153 patients, respectively.

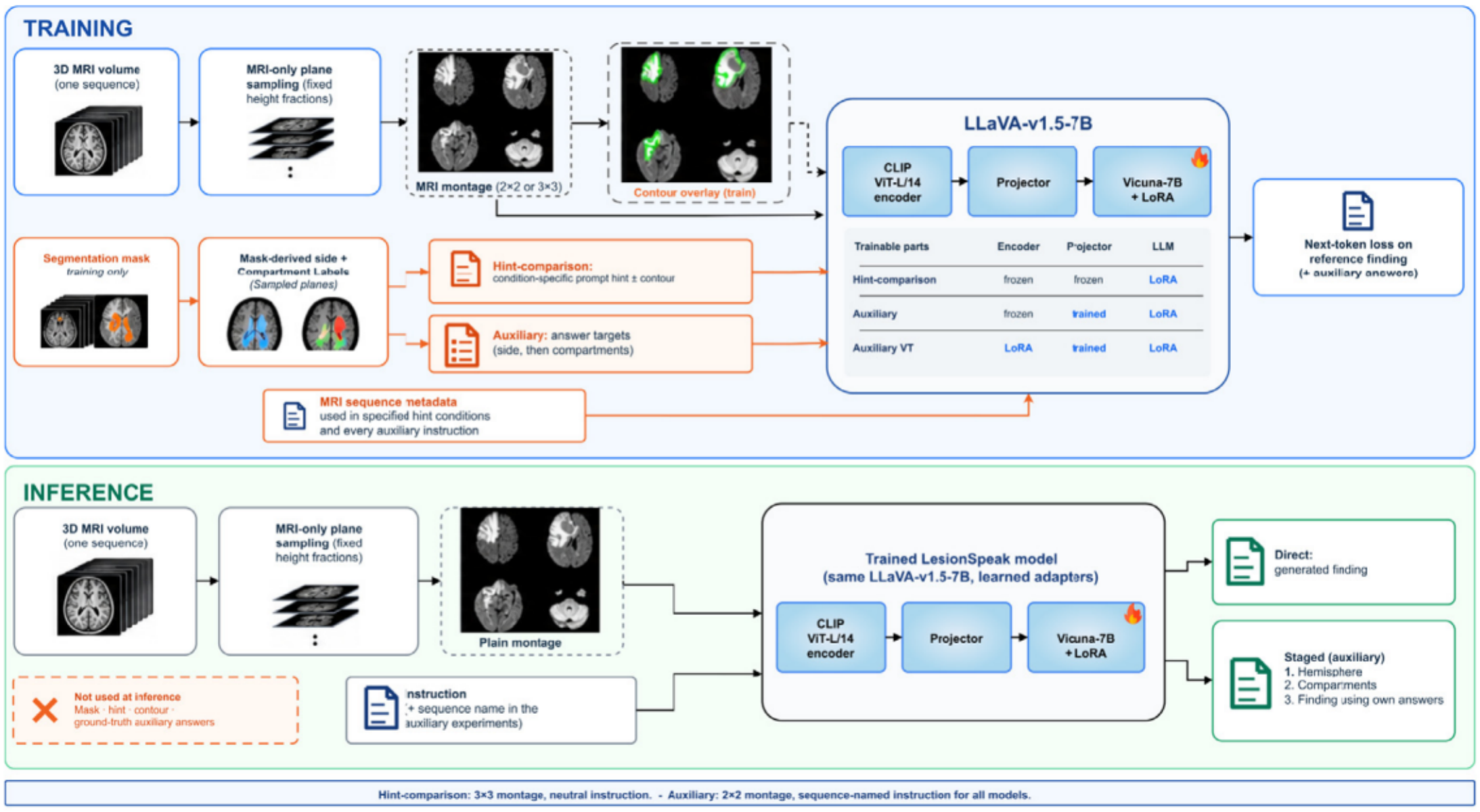

**Fig. 1.** Overview of the LesionSpeak training and inference framework.
Diagram of the LesionSpeak training and inference pipeline, showing MRI-only multislice input construction, segmentation-derived training supervision, acquisition-sequence metadata, model adaptation, and direct or staged finding generation.

**Table 1**
MRI-report records and patients by cohort and partition.

| Cohort | Training | Validation | Test |
|---|---|---|---|
| GLI | 644 (159) | 92 (23) | 184 (46) |
| MEN | 632 (157) | 100 (23) | 188 (46) |
| MET | 660 (165) | 92 (23) | 196 (49) |
| WMH | 2 (2) | 6 (6) | 12 (12) |
| Total | 1938 (483) | 290 (75) | 580 (153) |

Across these cohorts, records include native T1-weighted (T1n), post-contrast T1-weighted (T1c), T2-weighted (T2w), and T2 FLAIR MRI together with lesion segmentation labels. The target texts are sequence-specific findings from the AutoRG-Brain resource and were produced through GPT-4-assisted decomposition of radiologist-authored reports. They are therefore not independent radiologist annotations of each input montage. The test partition contains 142 T1n, 142 T1c, 142 T2w, and 154 FLAIR records. The dataset is lesion-enriched and does not contain a dedicated normal or lesion-negative cohort. Throughout the experiments, the generated text corresponds to a sequence-specific finding rather than a complete multi-sequence clinical report. We retain the term "report generation" when referring to the broader task family. Because a patient may contribute records from multiple MRI sequences, record counts exceed patient counts in each partition, as summarized in Table 1.

After defining the patient-level partitions, we next convert each 3D MRI volume into the 2D input format required by LLaVA. Each volume is represented as a fixed multislice montage. Volumes are canonicalized to RAS orientation using the NIfTI affine, with oblique volumes resampled to an axis-aligned grid. We then identify the inferior-superior extent of nonzero MRI voxels and sample planes at fixed relative positions within this extent. The hint-comparison experiments use nine planes at fractions 0.1–0.9 arranged as a 3 × 3 montage with 112-pixel tiles. The auxiliary experiments use four planes at fractions 0.2, 0.4, 0.6, and 0.8 arranged as a 2 × 2 montage with 168-pixel tiles. The 2 × 2 layout was selected using lesion coverage in the training set only; segmentation masks are not used to determine test input planes. Rendered target pixels occur on at least one sampled plane in 91.0% of training and 86.9% of validation records, although all records are retained. Thus, the montage is a sparse representation of the volume and may not capture every lesion finding. Each plane is displayed using a consistent radiological orientation, clipped to the 1st–99th percentiles of nonzero intensities, center-padded, and resized before montage construction. The final RGB input is 336 × 336. Algorithm 1 summarizes the preprocessing procedure.

### 3.2. Segmentation-derived lesion supervision

Segmentation-derived supervision is computed on the planes selected by the MRI-based preprocessing procedure. Segmentation masks are transformed to the canonical MRI grid using nearest-neighbor interpolation. Laterality is determined from lesion burden

**Algorithm 1:** MRI-based volume-to-montage preprocessing

**Input:** MRI volume $V$ with affine; fixed plane fractions $F$; montage layout
**Output:** MRI montage $I$ and selected-plane indices $Z$
Canonicalize $V$ to RAS orientation;
Resample oblique volumes to an axis-aligned MRI-derived grid when required;
Find the inferior-superior extent $[z_{\min}, z_{\max}]$ of nonzero MRI voxels;
Set $Z = \{z_{\min} + \lfloor f(z_{\max} - z_{\min}) \rfloor : f \in F\}$;
Order selected planes from superior to inferior;
**foreach** $z \in Z$ **do**
  Extract the MRI plane, apply the fixed radiological display orientation, clip and normalize nonzero intensities, center-pad, and resize to the fixed tile size;
Assemble the tiles into a 336 × 336 RGB montage $I$;
**return** $I$, $Z$

on the patient-left and patient-right sides of the canonical array midpoint. Records with no target voxels on the sampled planes receive a "no annotated target on sampled planes" label. Otherwise, a burden greater than 1.5 times the opposite side is labeled left- or right-predominant, while the remaining cases are labeled bilateral without hemispheric predominance. Among training records whose reference finding names exactly one side, the mask-derived laterality label agrees with the reference side in 1224 of 1274 cases (96.1%). MRI sequence identity is obtained independently from acquisition metadata and is not treated as a segmentation-derived attribute.

For the structured-hint conditions, the training tag has the form

$$[\texttt{LESION:}\mathit{lat};\ \mathit{compartments};\ \texttt{sequence:}\mathit{seq}]. \tag{1}$$

Here, laterality and compartment labels are derived from the segmentation mask, whereas sequence identity is acquisition metadata appended to the same structured prompt.

For GLI, segmentation labels 1, 2, and 3 represent necrotic tumor core, peritumoral edematous or infiltrated tissue, and enhancing tumor, respectively. MEN and MET use corresponding descriptors for the non-enhancing core, surrounding non-enhancing FLAIR hyperintensity, and enhancing tumor. WMH supervision uses label 1 for white matter hyperintensity. These labels are used differently across the two experiment families. In the hint-comparison conditions, lesion information is inserted into the training prompt. In the auxiliary conditions, mask-derived laterality and compartment information instead serves as explicit prediction targets before the reference finding. At test time, ground-truth lesion attributes are not supplied. Staged inference first generates the model's own attribute answers and then uses those answers when generating the finding. Algorithm 2 summarizes this training-time supervision procedure.

**Algorithm 2:** Training-time segmentation-derived supervision

**Input:** MRI geometry and selected planes $Z$; aligned segmentation $M$; cohort label dictionary; sequence identifier
**Output:** Laterality label, represented lesion compartments, and condition-specific training supervision
Transform $M$ to the canonical MRI grid using nearest-neighbor interpolation;
Restrict cohort-appropriate target labels to the selected planes $Z$;
Count target burdens $L$ and $R$ on the patient-left and patient-right sides of the canonical midpoint;
**if** $L + R = 0$ **then**
  set laterality to "no annotated target on sampled planes";
**else if** $L > 1.5R$ **then**
  set laterality to left-predominant;
**else if** $R > 1.5L$ **then**
  set laterality to right-predominant;
**else**
  set laterality to bilateral without hemispheric predominance;
List represented lesion compartments using the cohort-specific label dictionary;
Keep MRI sequence identity separate from segmentation-derived attributes;
Construct the condition-specific structured hint or auxiliary attribute targets;

### *3.3. Contour overlay augmentation*

To study whether spatially aligned contour supervision complements textual supervision, we create contour variants in which lesion boundaries from the segmentation mask are overlaid on the MRI-selected training planes. The binary lesion mask on each selected plane is resized to the tile size using nearest-neighbor interpolation, and a green contour formed by a one-pixel inner boundary followed by 3 × 3 dilation is drawn over the corresponding tile. Contours are used only during training; primary inference uses the corresponding plain MRI montage. We additionally evaluate a spatial control in which the contour is reflected horizontally within each tile while the MRI pixels, prompt, and reference finding remain unchanged. Fig. 2 illustrates the contour construction.

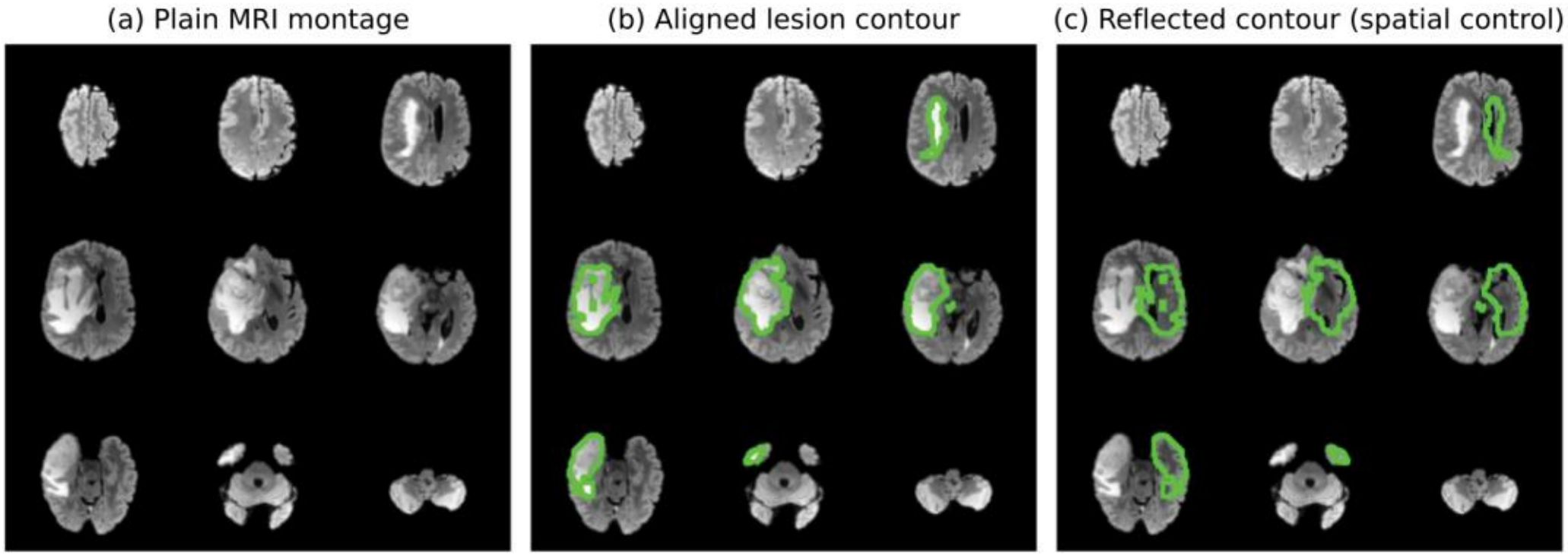

**Fig. 2.** Training-time contour augmentation and reflected-contour spatial control.
Comparison of a plain MRI montage, an anatomically aligned contour montage, and a spatially reflected contour montage.

### *3.4. LesionSpeak architecture*

LesionSpeak is implemented on top of LLaVA-v1.5-7B (Liu et al., 2024), which couples a CLIP ViT-L/14-336 visual encoder and multimodal projector to a Vicuna-7B decoder-only language model. Given an input image $I$ and instruction prompt $p$, the model generates a report $\hat{R} = (r_1, \ldots, r_T)$ according to

$$P_\theta(\hat{R} \mid I, p) = \prod_{t=1}^{T} P_\theta\left(r_t \mid r_{<t}, \phi_{\text{vis}}(I), \phi_{\text{txt}}(p)\right), \quad (2)$$

where $P_\theta(\cdot)$ denotes the conditional token distribution produced by the LoRA-adapted LLaVA model with parameters $\theta$, $\phi_{\text{vis}}$ denotes the visual encoder plus multimodal projector, and $\phi_{\text{txt}}$ denotes the text-side embedding and decoder stack.

In the hint-comparison experiments, the CLIP visual encoder and multimodal projector are frozen while LoRA updates are applied to the language decoder. In the auxiliary experiments, the multimodal projector is also trained. Variants denoted VT additionally apply LoRA to attention and feed-forward projections in the visual encoder.

### *3.5. LoRA fine-tuning strategy*

We adapt the pretrained backbone using LoRA (Hu et al., 2022), which parameterizes each weight update as

$$W' = W_0 + \Delta W = W_0 + \frac{\alpha}{r} BA, \quad (3)$$

where $W_0$ is the pretrained weight and $A$ and $B$ are trainable low-rank matrices. We use rank $r = 32$, scaling factor $\alpha = 64$, and LoRA dropout 0.05. Decoder LoRA is applied to the query, key, value, output, gate, up, and down projections and contains 79,953,920 trainable parameters. Training the multimodal projector adds 20,979,712 trainable parameters, giving 100,933,632 trainable parameters in auxiliary non-VT variants. VT variants additionally contain 14,155,776 vision-encoder LoRA parameters, for a total of 115,089,408 trainable parameters.

Experiments were conducted on a workstation equipped with two NVIDIA RTX 3090 GPUs (24 GB each), with each training run using a single GPU. We use AdamW with learning rate $1 \times 10^{-4}$, zero weight decay, cosine scheduling, warm-up ratio 0.03, and `bfloat16` precision without quantization. The per-device batch size is 1 with eight gradient-accumulation steps, giving an effective batch size of 8, and the context limit is 2048 tokens. The hint-comparison experiments use five training epochs, while the auxiliary experiments use three.

The training objective is token-level cross-entropy over supervised assistant responses:

$$\mathcal{L} = -\frac{1}{T}\sum_{t=1}^{T} \log P_\theta\left(y_t \mid y_{<t}, \phi_{\text{vis}}(I), \phi_{\text{txt}}(p)\right), \quad (4)$$

where $y_t$ denotes a supervised response token. In report-only variants, the supervised response is the reference finding. In auxiliary variants, supervision includes the lesion-attribute answers followed by the reference finding. Checkpoints are selected using the minimum teacher-forced mean loss on the validation set, with the earliest checkpoint retained in the event of a tie. To assess variation across training runs, the main hint-comparison conditions and Aux-LA-VT are trained with random seeds 42, 43, and 44. The evaluated seeds for the remaining auxiliary conditions are listed in Table 2.

**Table 2**
Model variants and training conditions.

| Model | Training input | Supervision target | Seeds |
|---|---|---|---|
| *Zero-shot baselines (validation)* | | | |
| LLaVA-v1.5/LLaVA-Med | Plain montage | None | – |
| *Hint-comparison experiments (3 × 3 montage)* | | | |
| LesionSpeak-FT | Plain montage; neutral prompt | Finding | 42, 43, 44 |
| LesionSpeak-Lat | Plain montage; laterality hint | Finding | 42, 43, 44 |
| LesionSpeak-Mod | Plain montage; sequence hint | Finding | 42, 43, 44 |
| LesionSpeak-LA | Plain montage; full structured hint | Finding | 42, 43, 44 |
| Contour-NoHint | Contour montage; neutral prompt | Finding | 42, 43, 44 |
| Contour-Mod | Contour montage; sequence hint | Finding | 42, 43, 44 |
| Reflected-Contour | Reflected contour; sequence hint | Finding | 42 |
| *Auxiliary experiments (2 × 2 montage; sequence-named prompt)* | | | |
| LesionSpeak-FT | Plain montage | Finding | 42 |
| LesionSpeak-FT-VT | Plain montage | Finding | 42 |
| LesionSpeak-Aux-LA | Plain montage | Side, compartments, finding | 42 |
| LesionSpeak-Aux-LA-VT | Plain montage | Side, compartments, finding | 42, 43, 44 |

### *3.6. Inference protocol and model variants*

At inference, models receive the plain MRI montage without a segmentation mask, contour, or ground-truth lesion attributes. The two experiment families use different montage, prompt, training, and inference configurations. Hint-comparison models use a neutral report-generation prompt in the primary condition, while all models in the auxiliary experiments receive MRI sequence identity in the base prompt. For staged auxiliary inference, the model first predicts lesion laterality and represented compartments and then generates the finding using its own predicted attributes. Ground-truth auxiliary answers are never inserted at test time. Adapted LLaVA-v1.5 models use the Vicuna-v1 conversation template and deterministic greedy decoding with `do_sample=False`, one beam, and at most 256 new tokens per turn. To quantify the effect of retaining privileged lesion information, we also evaluate a reference condition in which the full structured hint remains available to LesionSpeak-LA at inference.

We evaluate LLaVA-v1.5-7B and LLaVA-Med-v1.5-Mistral-7B (Li et al., 2023) as zero-shot baselines on the validation partition using neutral and sequence-named prompts. LLaVA-v1.5 uses the Vicuna-v1 template, whereas LLaVA-Med uses its native Mistral-instruct template. We also include an image-retrieval baseline that retrieves the nearest training image using frozen CLIP features and returns its associated report. Architecturally different systems such as AutoRG-Brain and BTReport are discussed as contextual related work because they were not evaluated under the same input protocol.

The adapted LesionSpeak conditions are summarized in Table 2. Because the hint-comparison and auxiliary experiments use different montage, prompt, and training configurations, LesionSpeak-FT denotes a separately trained protocol-matched baseline within each experiment family. Its scores should therefore be interpreted only within the corresponding protocol. VT denotes vision-encoder LoRA. The findings and auxiliary answers listed in the table are supervision targets rather than model inputs, and primary inference does not provide segmentation masks, contours, or ground-truth lesion attributes.

### *3.7. Evaluation metrics*

We evaluate generated reports using BLEU-1, BLEU-4, ROUGE-L, and BERTScore-F1 (Zhang et al., 2020). Corpus BLEU uses lowercase whitespace tokenization with NLTK method-1 smoothing. ROUGE-L is computed as mean F-measure using the stemmed `rouge_score` implementation. BERTScore-F1 uses `distilbert-base-uncased`, no inverse-document-frequency weighting, and baseline rescaling. Text-metric scores are reported on a 0–100 scale, and empty generations remain in the evaluation denominator.

To assess report grounding, we measure automated reference-text laterality agreement. A reference is eligible when a whole-word rule identifies exactly one of "left" or "right", and a prediction is counted as agreeing when the same rule returns the corresponding side without the opposite side. This yields 175 eligible validation records and 402 eligible test records from 129 patients. This endpoint measures agreement with the reference text and is not interpreted as clinical laterality accuracy.

Patient-image dependence is evaluated by replacing each MRI input with another patient's montage matched by cohort and MRI sequence while retaining the original prompt and reference finding. The donor image is drawn from a different patient. We report the difference between laterality agreement with the patient's own image and with the shuffled image. The image-retrieval baseline provides an additional comparison for understanding text-overlap scores that can be obtained by reusing an existing training report. We also report output length, BLEU brevity penalty, and the number of unique normalized reports to characterize templating and output diversity.

The primary auxiliary-model test endpoint is reference-text laterality agreement for LesionSpeak-Aux-LA-VT relative to LesionSpeak-FT. Checkpoints are selected using validation loss, and the test evaluation protocol is specified before test report generation. For conditions trained with multiple random seeds, we report the mean and sample standard deviation. Paired confidence intervals use 10,000 patient-clustered bootstrap samples, retaining all sequence records from each sampled patient. The contour-by-sequence-hint interaction is evaluated as (Contour-Mod−Contour-NoHint)−(Mod−FT). Exploratory ablation intervals are not adjusted for multiple comparisons. These automated measures assess text agreement and image dependence rather than clinical correctness.

**Table 3**
Report-generation performance on the auxiliary-protocol test partition.

| Method | BLEU-1 ↑ | BLEU-4 ↑ | ROUGE-L ↑ | BERT-F1 ↑ |
|---|---|---|---|---|
| LesionSpeak-FT | 27.80 | 8.44 | 32.72 | 56.90 |
| LesionSpeak-FT-VT | 27.98 | 8.59 | 32.72 | 56.95 |
| LesionSpeak-Aux-LA | 29.33 | 8.62 | 32.29 | 56.21 |
| LesionSpeak-Aux-LA-VT | 30.89 ± 1.16 | 10.46 ± 0.74 | 34.00 ± 0.95 | 57.27 ± 0.59 |

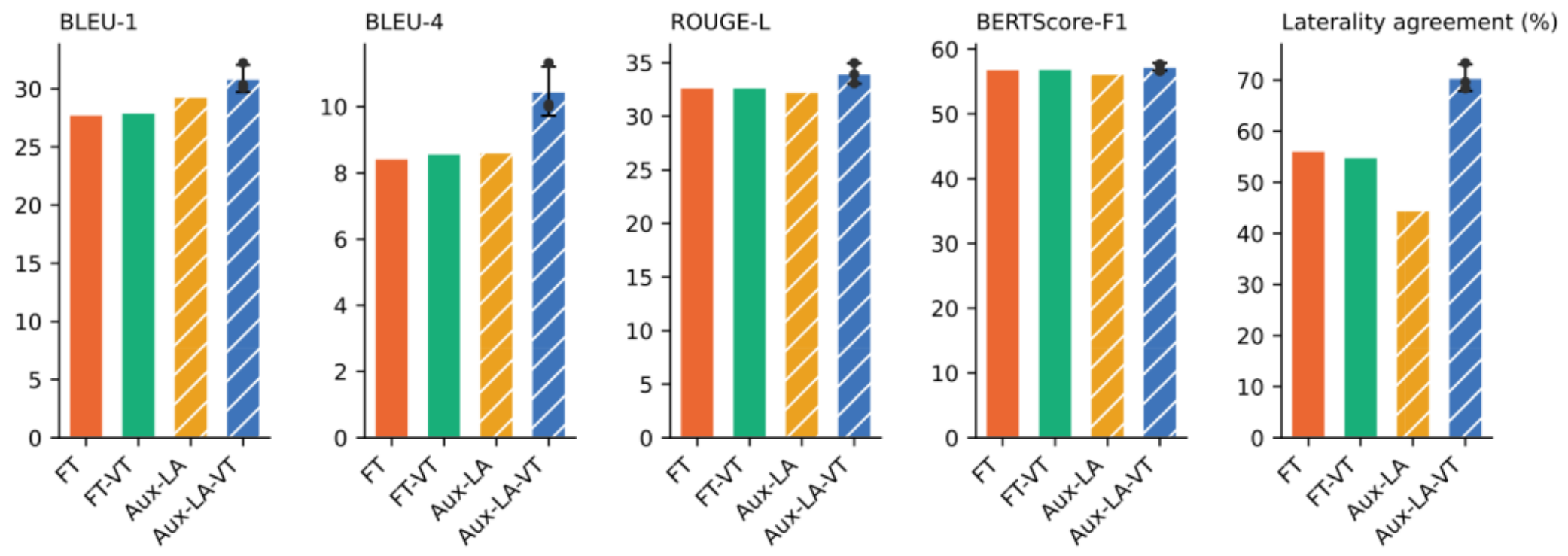

**Fig. 3.** Auxiliary-protocol test performance across text-similarity metrics and reference-text laterality agreement.
Bar charts of BLEU-1, BLEU-4, ROUGE-L, BERTScore-F1, and reference-text laterality agreement for LesionSpeak-FT, LesionSpeak-FT-VT, LesionSpeak-Aux-LA, and LesionSpeak-Aux-LA-VT on the test partition.

## 4. Experimental results

We organize the experimental results around three questions: (1) how the auxiliary lesion-aware vision-adapted model compares with fine-tuned controls, (2) whether generated reports show reference-consistent laterality and dependence on patient-specific MRI information, and (3) how structured hints, contour supervision, and individual model components affect report generation.

### *4.1. Main comparison with baselines*

We begin with the auxiliary-protocol comparison because it directly tests whether lesion-attribute supervision and visual adaptation improve finding generation under a common input setting. Table 3 and Fig. 3 summarize performance on the 580-record test partition from 153 patients. All models receive the same 2 × 2 MRI montage and MRI sequence metadata. FT denotes report-only fine-tuning, VT denotes vision-encoder LoRA, and Aux-LA denotes auxiliary lesion-attribute supervision. FT and FT-VT use direct generation, while Aux-LA and Aux-LA-VT use staged generation. LesionSpeak-FT, LesionSpeak-FT-VT, and LesionSpeak-Aux-LA are reported for seed 42. LesionSpeak-Aux-LA-VT is summarized across seeds 42, 43, and 44. In Fig. 3, individual points show seed-specific results and hatched bars identify staged inference. Reference-text laterality is evaluated on 402 eligible records and is examined separately in Section 4.2.

Before interpreting the adapted models, we place their text-overlap scores in context using validation-only baselines. With a neutral reporting prompt, zero-shot LLaVA-v1.5 reaches 10.86 BLEU-1 and 13.67 ROUGE-L, while LLaVA-Med reaches 18.22 and 16.65. Naming the MRI sequence in the instruction raises these values to 12.11 and 15.39 for LLaVA-v1.5 and to 24.55 and 24.20 for LLaVA-Med. Under the same hint-comparison validation protocol, LesionSpeak-FT reaches 24.78 ± 1.78 BLEU-1 and 29.68 ± 0.28 ROUGE-L across three training seeds.

Text overlap alone, however, does not establish that a generated finding reflects the patient's image. The image-retrieval baseline illustrates this limitation. Simply returning the report associated with the nearest training image produces 31.95 BLEU-1 and 28.62 ROUGE-L on validation. This motivates the later image-dependence analysis and provides context for interpreting the generative metrics in Table 3.

Within the auxiliary-protocol test comparison, LesionSpeak-Aux-LA-VT achieves BLEU-1 of 30.89 ± 1.16, BLEU-4 of 10.46 ± 0.74, ROUGE-L of 34.00 ± 0.95, and BERTScore-F1 of 57.27 ± 0.59 across three training seeds. Relative to the fixed LesionSpeak-FT seed-42 checkpoint, the ROUGE-L difference is +0.37 points for Aux-LA-VT seed 42, +2.26 for seed 43, and +1.21 for seed 44. The corresponding patient-clustered 95% confidence intervals are [−0.48, 1.21], [1.42, 3.11], and [0.53, 1.90]. Thus, the direction is positive across the three auxiliary-model seeds, although the interval for seed 42 includes zero.

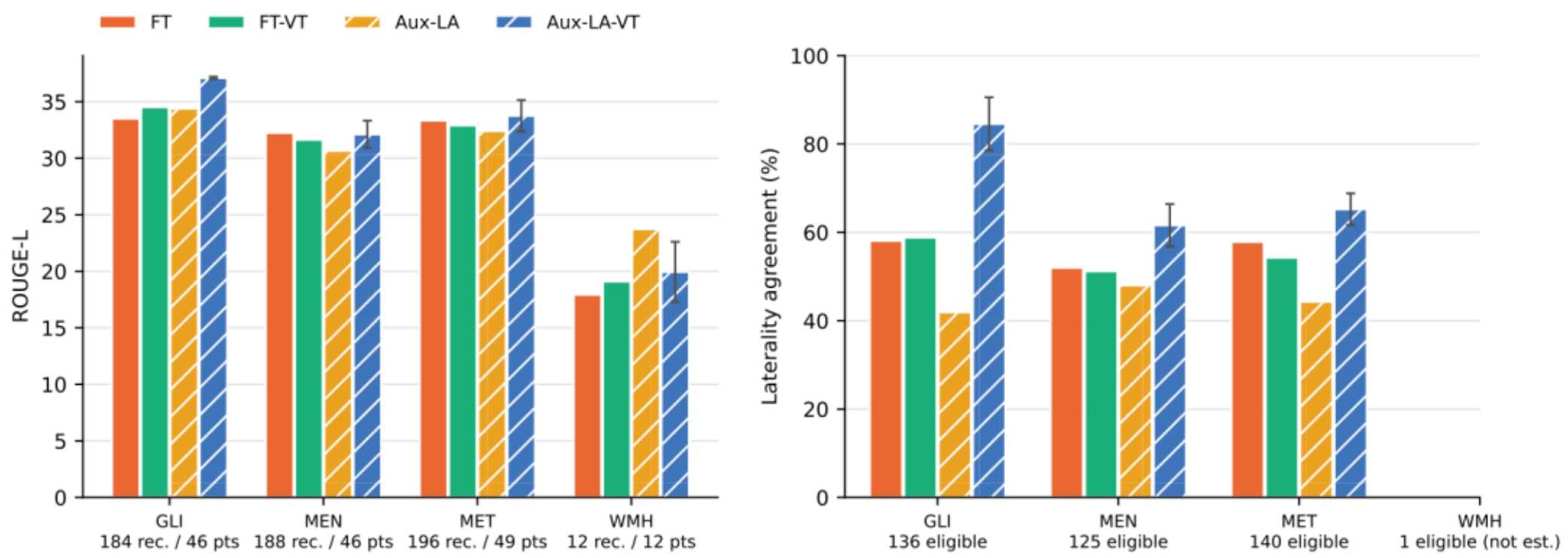

**Fig. 4.** Per-cohort test ROUGE-L and reference-text laterality agreement.
Per-cohort ROUGE-L and reference-text laterality agreement for LesionSpeak-FT, LesionSpeak-FT-VT, LesionSpeak-Aux-LA, and LesionSpeak-Aux-LA-VT on the test partition.

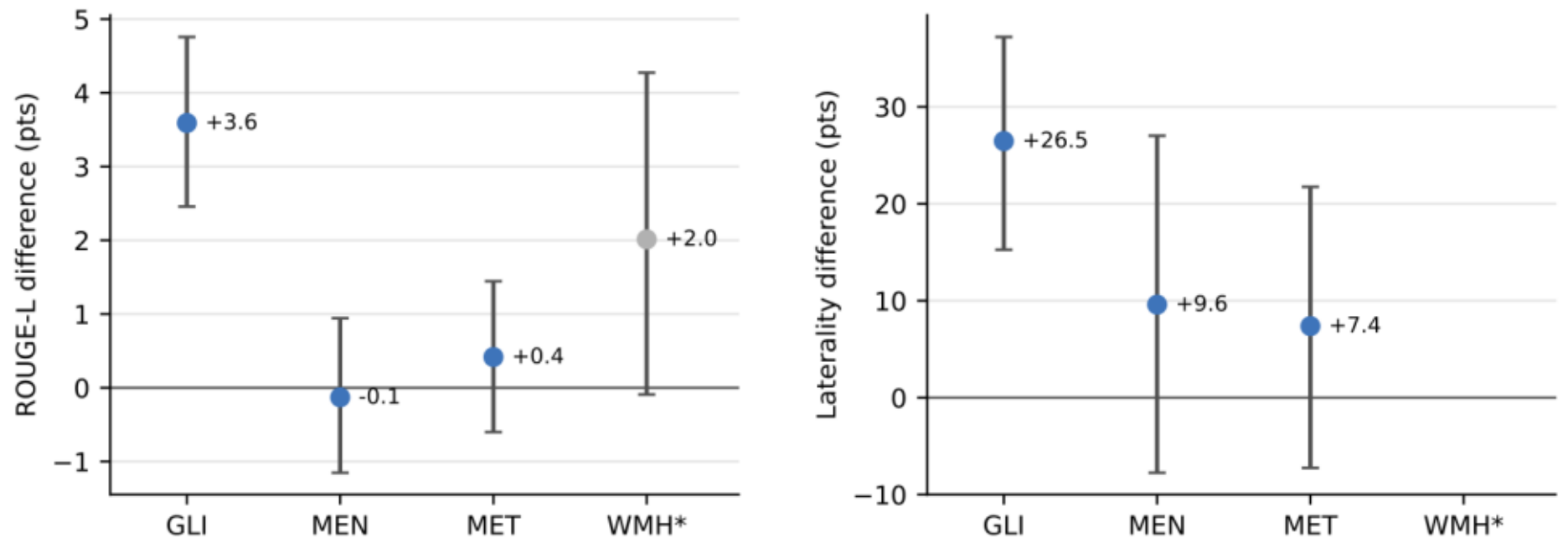

**Fig. 5.** Per-cohort Aux-LA-VT minus FT differences with patient-clustered 95% confidence intervals.
Per-cohort differences between LesionSpeak-Aux-LA-VT and LesionSpeak-FT in ROUGE-L and reference-text laterality agreement with patient-clustered confidence intervals.

The two component controls help interpret this result. Vision adaptation alone, represented by LesionSpeak-FT-VT, remains close to LesionSpeak-FT across the four text metrics. Auxiliary supervision with a frozen vision encoder, represented by LesionSpeak-Aux-LA, increases BLEU-1 but does not improve ROUGE-L or BERTScore-F1. These observations motivate the later ablation analysis, where we examine the supervision components and patient-image dependence more directly.

As a privileged reference condition, retaining the full segmentation-derived hint at inference gives LesionSpeak-LA 33.48 BLEU-1 and 35.50 ROUGE-L on validation (seed 42). Because this condition uses lesion information that is unavailable without a segmentation mask, it is treated only as a control for the effect of explicit hint access and not as a primary inference setting.

### *4.1.1. Per-Cohort analysis*

The aggregate comparison does not show whether the same pattern holds across pathology groups. We therefore examine ROUGE-L and reference-text laterality separately for GLI, MEN, MET, and WMH. Fig. 4 shows the cohort-specific scores, with Aux-LA-VT summarized across three seeds and the other models shown for seed 42. Fig. 5 then expresses the same comparison directly as Aux-LA-VT minus FT, using paired patient-clustered 95% confidence intervals. The clearest difference occurs in GLI, where ROUGE-L increases by 3.59 points ([2.46, 4.76]) and laterality agreement increases by 26.47 percentage points ([15.26, 37.21]). For MEN and MET, the intervals for both endpoints include zero. The figure labels show the corresponding record, patient, and laterality-eligible counts. The confidence intervals reflect patient sampling conditional on the evaluated checkpoints and do not include training-seed uncertainty.

WMH cannot be interpreted on the same footing as the three tumor cohorts because it is sparsely represented. The partition contains only 2 WMH training records and 12 test records, compared with 632–660 training records and 184–196 test records for each tumor cohort. Only one WMH test record is eligible for the reference-text laterality endpoint, so WMH laterality is not estimated.

## *4.2. Reference laterality and image dependence*

The preceding text-similarity results show how closely generated findings resemble the references, but they do not determine whether the model's laterality statement depends on the patient's MRI. We therefore evaluate reference-text laterality agreement together with an image-replacement control. A reference is included when it names exactly one side, yielding 402 eligible test records from 129 patients. For the image-dependence analysis, we replace each MRI with another patient's montage from the same cohort and sequence while retaining the original prompt and reference finding. Table 4 reports laterality agreement on the 402 eligible records, the own-minus-shuffled difference, and the number of unique normalized findings across all 580 test records. Brackets show paired patient-clustered 95% confidence intervals. Fig. 6 visualizes the own-image and shuffled-image comparison. Fig. 6 labels each own-minus-shuffled difference with its paired patient-clustered 95% confidence interval.

**Table 4**
Reference-text laterality, image dependence, and output diversity on the test partition.

| Method | Laterality (%) | Own−shuffled (pts) | Unique reports |
|---|---|---|---|
| LesionSpeak-FT | 56.22 | 0.00 [−3.21, 3.22] | 52 |
| LesionSpeak-FT-VT | 54.98 | – | 75 |
| LesionSpeak-Aux-LA | 44.53 | – | 99 |
| LesionSpeak-Aux-LA-VT, seed 42 | 69.65 | +22.64 [14.36, 30.79] | 178 |
| LesionSpeak-Aux-LA-VT, seed 43 | 73.38 | +24.13 [16.05, 32.04] | 236 |
| LesionSpeak-Aux-LA-VT, seed 44 | 68.41 | +16.92 [9.52, 24.28] | 211 |

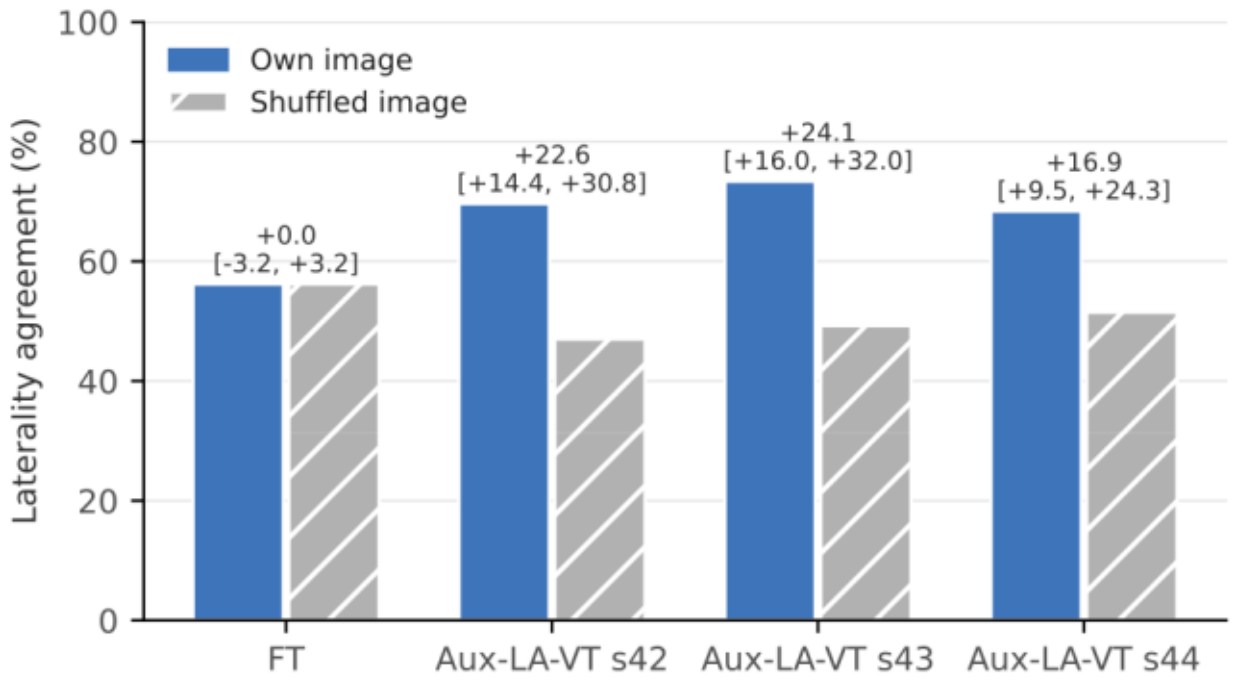

**Fig. 6.** Reference-text laterality agreement with own and shuffled MRI inputs.
Reference-text laterality agreement with the patient's own MRI and with a shuffled MRI for LesionSpeak-FT and three LesionSpeak-Aux-LA-VT seeds.

LesionSpeak-Aux-LA-VT reaches mean reference-text laterality agreement of 70.48 $\pm$ 2.59% across three training seeds, compared with 56.22% for the protocol-matched LesionSpeak-FT seed-42 checkpoint. Relative to that FT checkpoint, the laterality differences are +13.43 percentage points for Aux-LA-VT seed 42, +17.16 for seed 43, and +12.19 for seed 44. The corresponding patient-clustered 95% confidence intervals are [0.72, 25.96], [7.52, 26.90], and [5.21, 19.20]. These comparisons remain conditional on the single evaluated FT training seed.

The image-replacement control provides the more direct test of patient-image dependence. Replacing the patient's MRI reduces laterality agreement by 22.64 percentage points for Aux-LA-VT seed 42, 24.13 points for seed 43, and 16.92 points for seed 44. All three patient-clustered confidence intervals remain above zero. The corresponding difference for LesionSpeak-FT seed 42 is 0.00 points with a confidence interval of [−3.21, 3.22]. Taken together, these results support dependence of the auxiliary model's reference-text laterality endpoint on patient-specific MRI information. They do not establish the clinical correctness of the complete generated finding.

### *4.3. Ablation study*

The main comparison suggests that lesion supervision is not uniformly beneficial, so we next examine which forms of supervision account for this behavior. We begin with the matched hint-comparison experiments, which vary laterality hints, sequence hints, full structured hints, and training-time contours while keeping preprocessing and primary inference settings fixed. Table 5 and Fig. 7 summarize the 580-record test results across seeds 42, 43, and 44. All six primary conditions use the same 3 × 3 MRI montage and neutral inference prompt, with no hint or contour supplied at inference. The reflected-contour control is evaluated for seed 42 only. Table 5 reports metric means and sample standard deviations, while the distinct-output column lists counts for seeds 42, 43, and 44 in that order. In Fig. 7, bars show the three-seed means, error bars show sample standard deviations, individual points show seed-specific results, and the dashed line marks the LesionSpeak-FT mean.

Across these matched conditions, LesionSpeak-FT has the highest mean ROUGE-L, while LesionSpeak-Mod and LesionSpeak-Contour-Mod show substantial variation across training seeds. The full-hint condition is especially sensitive to whether the hint remains available. On validation, LesionSpeak-LA reaches 14.17 $\pm$ 7.37 BLEU-1 and 18.67 $\pm$ 3.40 ROUGE-L when the training-time hint is removed. Retaining the full hint increases these values to 29.04 $\pm$ 1.29 and 34.95 $\pm$ 0.61 across seeds 42, 43, and 44. This confirms that inference-time hint availability materially changes report generation, but it does not establish how that information is represented internally.

**Table 5**

Hint-comparison test results: mean ± sample standard deviation across seeds 42, 43, and 44. The reflected-contour control uses seed 42 only. Distinct-output counts follow seed order 42/43/44.

| Model | BLEU-1 | BLEU-4 | ROUGE-L | BERT-F1 | Distinct |
|---|---|---|---|---|---|
| LesionSpeak-FT | 24.07 ± 1.50 | 7.12 ± 0.66 | 30.43 ± 0.16 | 53.17 ± 0.35 | 37/63/44 |
| LesionSpeak-Lat | 24.32 ± 3.52 | 6.74 ± 1.07 | 28.62 ± 1.56 | 51.50 ± 1.93 | 33/44/68 |
| LesionSpeak-Mod | 20.45 ± 9.00 | 5.06 ± 2.51 | 22.69 ± 6.70 | 41.97 ± 9.55 | 16/32/20 |
| LesionSpeak-LA | 13.52 ± 7.21 | 2.74 ± 1.75 | 18.83 ± 3.38 | 36.63 ± 5.37 | 1/9/16 |
| LesionSpeak-Contour-NoHint | 20.07 ± 5.46 | 5.50 ± 0.98 | 28.23 ± 1.28 | 53.44 ± 1.20 | 25/70/13 |
| LesionSpeak-Contour-Mod | 22.38 ± 11.98 | 4.71 ± 2.74 | 23.63 ± 6.84 | 44.21 ± 12.74 | 50/19/11 |
| Contour-Mod, reflected contour | 15.76 | 3.53 | 18.62 | 35.64 | 32 |

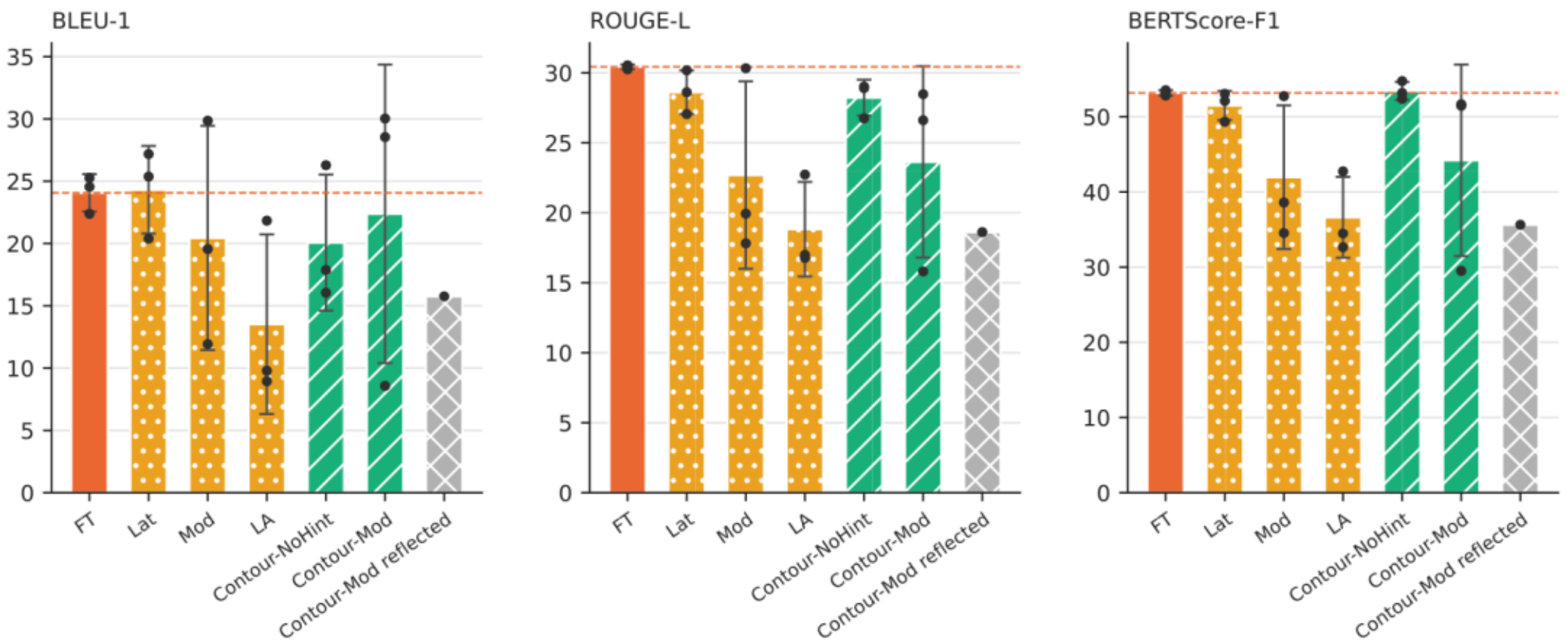

**Fig. 7.** Three-seed hint-comparison performance on the test partition.
Test BLEU-1, ROUGE-L, and BERTScore-F1 for the six hint-comparison conditions and the reflected-contour control, with individual training seeds shown.

Because contour supervision and sequence hints can occur independently, we next examine their interaction using the four matched conditions LesionSpeak-FT, LesionSpeak-Mod, LesionSpeak-Contour-NoHint, and LesionSpeak-Contour-Mod. Their mean test ROUGE-L values form the 2 × 2 comparison shown below.

| | No sequence hint | Sequence hint |
|---|---|---|
| No contour | LesionSpeak-FT: 30.43 ± 0.16 | LesionSpeak-Mod: 22.69 ± 6.70 |
| Contour | Contour-NoHint: 28.23 ± 1.28 | LesionSpeak-Contour-Mod: 23.63 ± 6.84 |

The contour-by-sequence-hint interaction on ROUGE-L is +10.46 points for seed 42 (95% patient-clustered confidence interval [9.15, 11.77]), −0.77 for seed 43 ([−1.74, 0.17]), and −0.24 for seed 44 ([−1.42, 0.94]). Thus, a positive interaction is clearly resolved only for seed 42. LesionSpeak-Contour-Mod remains below the matched LesionSpeak-FT model in ROUGE-L for all three seeds, by −3.92, −14.44, and −2.04 points, respectively.

As a spatial control, the correctly aligned contour exceeds the horizontally reflected contour by 8.00 ROUGE-L points in seed 42 (95% confidence interval [6.92, 9.07]). This single-seed result shows sensitivity to contour alignment in that experiment, but does not establish a seed-robust contour benefit over plain fine-tuning.

The text-metric ablations show how the supervision conditions affect report similarity, but they still do not reveal whether laterality depends on the patient's image. We therefore apply the image-replacement control to the hint-comparison runs and to the auxiliary component ablations. Table 6 and Fig. 8 summarize this analysis on the 175 laterality-eligible validation records. Own−shuffled denotes the change in reference-text laterality agreement after replacing the MRI with another patient's montage from the same cohort and sequence. For models with an intermediate hemisphere prediction, the side-answer column reports accuracy against the corresponding auxiliary side target for the patient's own and shuffled images. Brackets in Table 6 show paired patient-clustered 95% confidence intervals.

None of the 18 hint-comparison runs show a resolved positive own-minus-shuffled laterality effect. In contrast, LesionSpeak-Aux-LA-VT shows differences of +16.6, +16.6, and +17.7 percentage points across its three evaluated seeds, with all three confidence intervals above zero. The component controls do not show the same resolved effect. Side-only supervision, compartment-only supervision, shuffled side labels, report-derived side labels, vision adaptation without auxiliary supervision, and auxiliary supervision with a frozen vision encoder all have confidence intervals that include zero. These controls associate the observed image-dependence result with the combined side-and-compartment formulation and vision adaptation under the evaluated configuration, without establishing that any single component is universally necessary.

**Table 6**
Validation image-dependence results for the auxiliary ablations.

| Auxiliary supervision | Seed | Own−shuffled (pts) | Side answer (%) |
|---|---|---|---|
| Side + compartments (Aux-LA-VT) | 42 | +16.6 [7.1, 26.8] | 79.7/58.2 |
| | 43 | +16.6 [4.9, 28.3] | 75.9/54.4 |
| | 44 | +17.7 [5.9, 30.1] | 55.3/35.9 |
| Side only | 42, 43 | +8.0 [−4.2, 20.7]; +0.0 [−8.9, 8.6] | 62.9/56.5; 47.7/46.4 |
| Compartments only | 42 | −1.1 [−8.6, 6.2] | – |
| Side labels shuffled across patients | 42 | +0.0 [−3.8, 4.0] | 52.3/53.6 |
| Side labels from report text | 42, 43 | −6.9 [−15.3, 1.7]; +4.0 [−3.0, 11.8] | 47.3/56.1; 46.4/42.2 |
| None, vision LoRA only (FT-VT) | 42 | −2.9 [−8.1, 2.5] | – |
| Side + compartments, frozen encoder (Aux-LA) | 42 | −2.9 [−13.2, 6.8] | 51.9/54.4 |

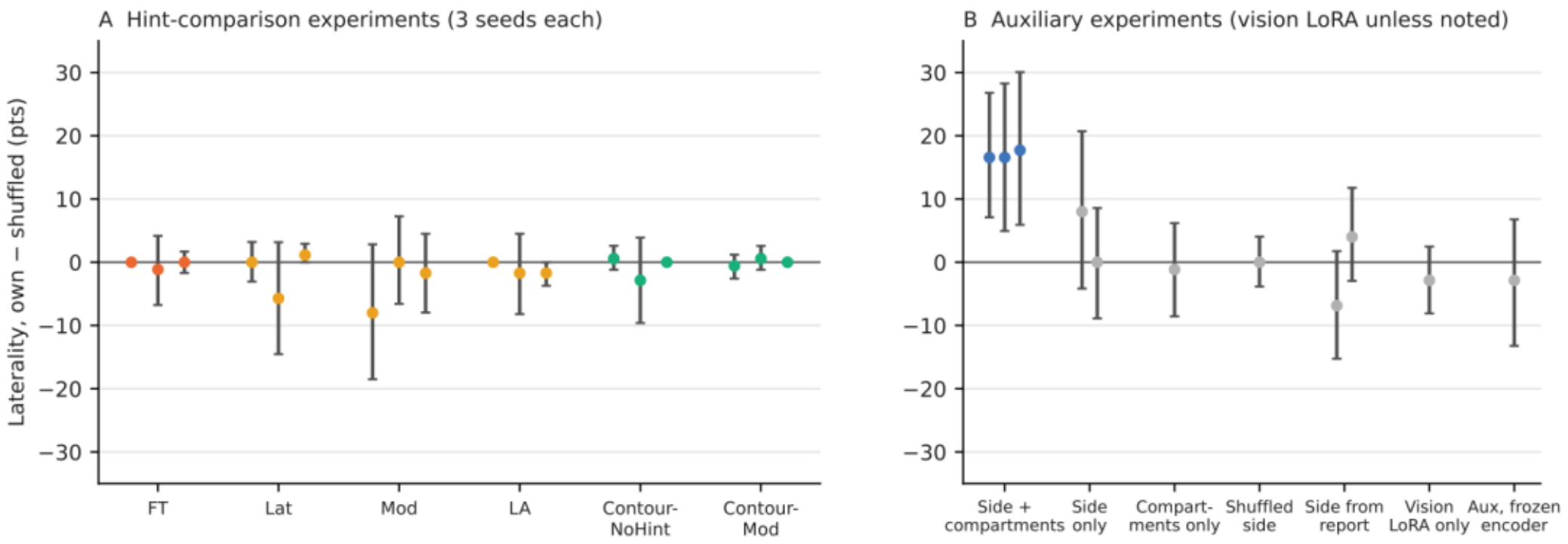

**Fig. 8.** Validation image-dependence ablations for the hint-comparison and auxiliary experiments.
Own-minus-shuffled reference-text laterality differences with confidence intervals for the hint-comparison runs and auxiliary ablations.

A separate contour-augmented auxiliary control provides another caution about training-time cues. When its training contour is removed, the single-seed model answers ''not visible'' for 274 of 290 plain validation montages. This behavior indicates strong sensitivity to contour availability in that run and reinforces the distinction between learning from a training cue and relying on it.

Fig. 8 visualizes the same analysis. Panel A shows the hint-comparison runs and Panel B shows the auxiliary component controls. Points represent own-minus-shuffled laterality differences, with paired patient-clustered 95% confidence intervals shown as error bars.

### *4.4. Output diversity and qualitative analysis*

The quantitative analyses also reveal substantial differences in output diversity. LesionSpeak-FT seed 42 produces 52 unique normalized findings among 580 test records, corresponding to 8.97% unique outputs, and its most frequent finding occurs 114 times. LesionSpeak-Aux-LA-VT produces 178, 236, and 211 unique findings across seeds 42, 43, and 44, with the most frequent finding occurring 49, 31, and 42 times. Mean output length remains similar, at 22.70 words for LesionSpeak-FT and 23.77, 24.98, and 23.55 words for Aux-LA-VT. The corresponding BLEU brevity penalties are 0.638, 0.681, 0.729, and 0.673. Thus, the auxiliary model produces more varied outputs without a large change in average length, although greater variation alone does not establish factual correctness.

To complement these aggregate diagnostics, Fig. 9 shows three representative test cases and compares the reference finding with outputs from LesionSpeak-FT, LesionSpeak-FT-VT, LesionSpeak-Aux-LA, and LesionSpeak-Aux-LA-VT using seed 42. Cases (a) and (b) are the first GLI and MET records by identifier for which Aux-LA-VT states the reference side and FT does not. Case (c) is the first eligible record for which Aux-LA-VT states the opposite side. Green and red indicate agreement and disagreement with reference-text laterality, and the staged auxiliary outputs also show the model's intermediate hemisphere answer.

In cases (a) and (b), LesionSpeak-FT and LesionSpeak-FT-VT name the hemisphere opposite to the reference, whereas LesionSpeak-Aux-LA-VT produces the reference-consistent side and its generated finding follows its intermediate hemisphere answer. In case (b), LesionSpeak-Aux-LA also produces the reference-consistent side but places the lesion in the right frontal lobe rather than the right cerebellum, illustrating that laterality agreement does not imply correct regional localization. Case (a) shows another limitation: LesionSpeak-Aux-LA-VT introduces a lesion size that is not stated in the reference. Case (c) shows the opposite behavior. Both auxiliary models predict ''right'' for a reference finding that states a left frontal lesion, and their generated findings carry that laterality into the report, while LesionSpeak-FT names the reference-consistent side. These examples illustrate how the intermediate hemisphere answer can influence the laterality expressed in staged report generation and how an incorrect intermediate answer can propagate to the generated finding. They are qualitative examples and are not used as evidence of clinical accuracy or patient-image dependence.

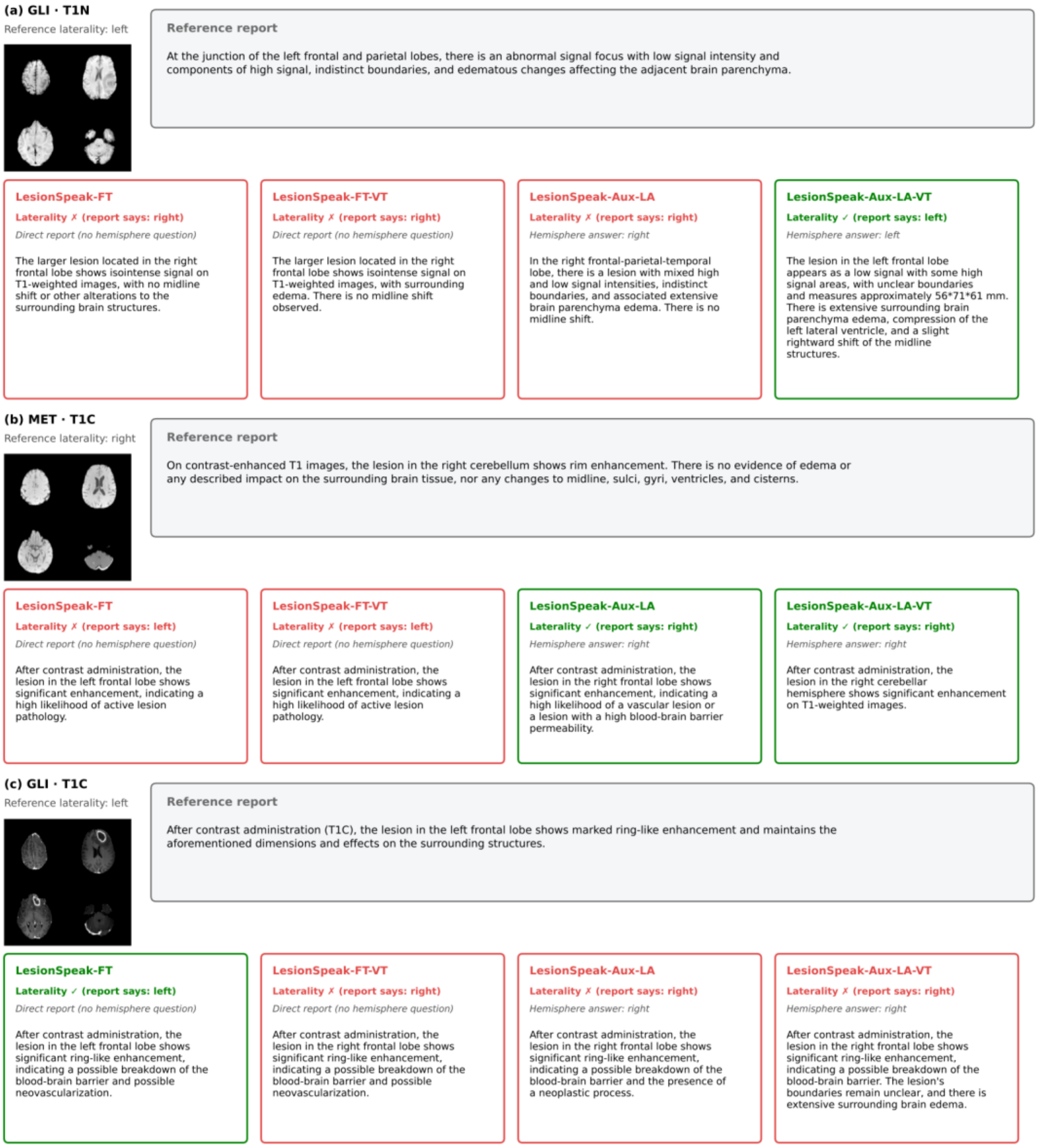

**Fig. 9.** Qualitative comparison of three selected test cases across the four main auxiliary-protocol models.
Three test cases showing the MRI montage, reference finding, and generated findings from LesionSpeak-FT, LesionSpeak-FT-VT, LesionSpeak-Aux-LA, and LesionSpeak-Aux-LA-VT, with reference-text laterality agreement indicated for each output.

## 5. Discussion

The central finding of this study is that training-time lesion supervision does not by itself ensure that generated findings depend on patient-specific MRI information. In the hint-comparison experiments, laterality, sequence, full-hint, and contour-based supervision do not show a consistent text-metric advantage over plain fine-tuning across training seeds, and none of the evaluated hint-comparison runs shows a resolved positive own-minus-shuffled laterality effect. By contrast, the auxiliary lesion-aware model that combines mask-derived side and compartment supervision with vision adaptation shows patient-image dependence of reference-text laterality agreement across three training seeds. The component controls sharpen this interpretation: vision adaptation without auxiliary lesion supervision, auxiliary supervision with a frozen vision encoder, side-only or compartment-only supervision, shuffled side labels, and report-derived side labels do not show a resolved positive own-minus-shuffled laterality effect under their evaluated configurations. These results indicate that the form of supervision and the pathway being adapted both matter, while avoiding the stronger conclusion that any single component independently guarantees visual grounding.

This finding also clarifies the role of segmentation masks in the framework. Segmentation masks can provide training-time supervision without being required to construct the test input or generate the finding. Test montages are sampled from MRI intensities using fixed relative height fractions rather than lesion masks, and segmentation-derived laterality and compartment information is not supplied to the model at inference. In the staged auxiliary setting, the model generates its own lesion-attribute answers before producing the finding rather than receiving ground-truth mask-derived answers. This separation allows lesion annotations available during model development to contribute supervision without assuming that the same annotations will be available when the trained model is used.

Separating training-time supervision from inference-time input also requires care in interpreting what the grounding metrics measure. MRI sequence identity is acquisition metadata rather than a lesion attribute derived from segmentation, and it is supplied to all compared models in the auxiliary experiments. Consequently, sequence-specific language in these experiments cannot be interpreted as evidence that the model inferred sequence identity from image appearance. Similarly, reference-text laterality agreement measures whether the generated finding states the same hemisphere as the reference, but does not establish finer regional localization or complete anatomical correctness. The qualitative examples illustrate this distinction: a generated finding can agree with the reference laterality while naming an incorrect anatomical region. Patient-image replacement therefore provides an important complementary control because it tests whether the measured laterality endpoint changes with the patient's MRI rather than relying on report similarity alone.

The contour experiments reinforce the same caution from a different direction. Although the correctly aligned contour outper-forms a horizontally reflected contour in the seed-42 spatial control, the contour-by-sequence-hint interaction varies substantially across training seeds. In the separate contour-augmented auxiliary control, the model answers ''not visible'' for 274 of 290 plain validation montages when the training contour is absent, indicating strong sensitivity to contour availability in that run rather than a general contour benefit. The aggregate laterality difference is also not uniform across pathology groups: the clearest cohort-level difference occurs in GLI, whereas the MEN and MET confidence intervals include zero. Taken together, these analyses support evaluating text similarity, reference agreement, and patient-image dependence as distinct properties rather than using any single metric as evidence of overall report correctness.

## 6. Limitations and future work

The current study provides evidence about text agreement and patient-image dependence, but several factors limit how broadly the present findings can be interpreted. Clinical interpretation of the findings was informed by input from a radiologist co-author. However, no blinded reader study or independent clinical evaluation of the generated findings was performed, and the reported text-similarity, reference-laterality, and image-shuffling measures should not be interpreted as measures of clinical correctness. The reference targets are sequence-specific findings produced through GPT-4-assisted decomposition of radiologist-authored reports rather than through independent annotations of the selected MRI montages. In addition, laterality agreement captures hemispheric consistency but not finer regional localization. Future evaluation should therefore include blinded radiologist assessment of factual correctness, lesion localization, unsupported findings, and clinical usefulness.

The framework also represents each 3D MRI volume using a sparse multislice montage rather than the complete examination. Although fixed plane sampling removes test-mask dependence, relevant lesions may fall outside the selected planes. The auxiliary $2 \times 2$ layout includes a rendered target on at least one sampled plane for 91.0% of training records and 86.9% of validation records, while retaining all records. Future work should investigate denser multislice inputs and volumetric vision-language models, such as Med3DVLM (Xin et al., 2026), to preserve more complete 3D context.

Generalization is further limited by the dataset composition. The corpus is lesion-enriched, does not contain a dedicated normal or lesion-negative cohort, and is derived entirely from AutoRG-Brain data. WMH is particularly underrepresented, with only 2 training records and 12 test records. To assess broader generalization, we need evaluation on lesion-negative examinations, better-represented pathology groups, and external datasets from different institutions and acquisition settings.

Some comparisons also have unequal training-seed coverage, and several auxiliary ablations are evaluated with only one or two seeds. The reported patient-clustered confidence intervals capture sampling variation across patients but not training-seed uncertainty. In addition, direct and staged generation differ in inference procedure. The hint-comparison and auxiliary experiment families additionally differ in montage layout, sequence-metadata availability, projector training, vision adaptation, and training duration, so comparisons between these families do not isolate a single causal factor. The auxiliary test analysis should therefore be interpreted as exploratory rather than as an independent confirmatory evaluation. Future studies should use balanced multi-seed evaluations and more controlled experiments that vary one component at a time. Evaluating the same approach with additional medical vision-language backbones will also help determine whether the observed image-dependence behavior extends beyond LLaVA-v1.5-7B.

### *6.1. Conclusion*

This work presented LesionSpeak, a segmentation-guided LoRA adaptation framework for brain MRI report generation that uses lesion annotations for training supervision without requiring segmentation masks at inference. The controlled hint-comparison experiments show substantial variation across supervision conditions and training seeds and do not support a general advantage of partial hints or contour-plus-sequence supervision over plain fine-tuning. In the auxiliary extension, combining mask-derived side and lesion-compartment supervision with vision adaptation and staged inference yields higher reference-text laterality agreement and a reduction in that agreement when the patient MRI is replaced. This offers a practical approach to using segmentation-derived supervision during training while preserving inference without segmentation masks. Additionally, the evaluated component controls, including side-only, compartment-only, and frozen-encoder auxiliary variants, do not show a resolved positive own-minus-shuffled laterality effect on validation. These results provide evidence that the measured laterality endpoint depends on patient-specific MRI information, while text similarity, reference agreement, output diversity, and image dependence remain distinct from clinical correctness. Overall, LesionSpeak demonstrates a practical approach for using segmentation-derived supervision during training while preserving inference without segmentation masks and motivates further evaluation with volumetric inputs, external and lesion-negative cohorts, additional model families, and clinical assessment.

## CRediT authorship contribution statement

**Almustapha A. Wakili:** Ideation, conceptualization, methodology, investigation, experimental design and execution, validation, evaluation, and writing of the original draft. **Muhammad Adam:** Review, comments, and manuscript editing. **Julius Akan:** Review, comments, and manuscript editing. **Stephanie Y. Jo:** Clinical review, assessment of clinical relevance, interpretation of findings, radiology terminology and workflow guidance, and review and editing of the manuscript. **Lei Zhang:** Idea validation, project administration, supervision, and review and editing of the manuscript. **Woosub Jung:** Idea validation, project administration, supervision, and review and editing of the manuscript.

## Code availability

Source code is publicly available at https://github.com/Binwakil/LesionSpeak. The repository includes MRI preprocessing, training and generation implementations for the hint-based and auxiliary-supervision protocols, evaluation utilities, configuration files, dependency specifications, and synthetic tests. Project code is distributed under the Apache License 2.0, while vendored components retain their respective licenses and attribution. Dataset and pretrained-model terms apply separately. Trained study checkpoints and study-specific test-generation queues are not included. The documented generation workflow operates on validation data, and exact reproduction of the reported test results requires the corresponding study-specific artifacts.

## Ethics statement

This study uses existing de-identified MRI data and associated annotations from the RadGenome-Brain MRI resource and involves no new participant recruitment or data collection. Institutional review board approval was not required for this secondary analysis of existing data. All data were accessed and used in accordance with the applicable source-dataset terms and data-governance requirements.

## Funding

This research was supported by a School of Emerging Technologies (SET) Seed Grant from Towson University.

## Declaration of competing interest

The authors declare that they have no known competing financial interests or personal relationships that could have appeared to influence the work reported in this paper.

## Data availability

This study uses MRI volumes, segmentation annotations, and sequence-specific findings from the RadGenome-Brain MRI resource associated with AutoRG-Brain (Lei et al., 2024). Source access information is available at https://github.com/ljy19970415/AutoRG-Brain. Access and reuse are subject to the applicable source-dataset terms. The LesionSpeak repository does not redistribute the underlying data or the study-specific patient split identifiers. It documents the required input format and provides checks for patient overlap across user-supplied partitions.